\documentclass[11pt,a4paper]{article}
\usepackage{xcolor}
\usepackage{authblk}
\usepackage[hyperref]{emnlp-ijcnlp-2019}
\usepackage{times}
\usepackage{latexsym}
\usepackage{mathtools}
\usepackage{dsfont}
\usepackage{amsmath, amsthm, amssymb, enumitem}
\usepackage{hyperref}
\usepackage{graphicx}
\usepackage{lipsum}
\usepackage{bm}
\usepackage{url}
\usepackage{multirow}
\usepackage{anyfontsize}
\usepackage{tabu}
\usepackage{booktabs}

\DeclarePairedDelimiterX{\infdivx}[2]{(}{)}{%
  #1\;\delimsize\|\;#2%
}
\newcommand{\infdiv}{{KL}\infdivx}

\usepackage{array}
\newcolumntype{P}[1]{>{\centering\arraybackslash}p{#1}}
\newcolumntype{M}[1]{>{\centering\arraybackslash}m{#1}}
\newcolumntype{C}{@{\hskip3.5pt}c@{\hskip3.5pt}}
\newcolumntype{T}{@{\hskip3pt}c@{\hskip3pt}}

\newcommand\Tstrut{\rule{0pt}{2.0ex}}       % "top" strut
\newcommand\Bstrut{\rule[-1.2ex]{0pt}{0pt}} % "bottom" strut
\aclfinalcopy % Uncomment this line for the final submission

\author[ \hspace{0.7mm}1]{Seanie Lee\footnote[1]{Equal contribution}}
\newcommand\CoAuthorMark{\footnotemark[\arabic{footnote}]} % get the current value
\author[ \hspace{0.7mm}1]{Donggyu Kim\protect\CoAuthorMark\thanks{\hspace{1.5mm}Equal contribution}}
\author[ \hspace{0.9mm}2]{Jangwon Park\footnote[1]{Equal contribution}}
\affil[1]{42Maru, Seoul, Korea}
\affil[2]{Samsung Research, Seoul, Korea}
\affil[ ]{\tt {\{lsnfamily02,donggyukimc\}@42maru.ai}, jang1.park@samsung.com}

\date{}

\begin{document}
\title{Domain-agnostic Question-Answering with Adversarial Training}
\maketitle
\begin{abstract}
  Adapting models to new domain without fine-tuning is a challenging problem in deep learning. In this paper, we utilize an adversarial training framework for domain generalization in Question Answering (QA) task. Our model consists of a conventional QA model and a discriminator. The training is performed in the adversarial manner, where the two models constantly compete, so that QA model can learn domain-invariant features. We apply this approach in MRQA Shared Task 2019 and show better performance compared to the baseline model.
\end{abstract}

\section{Introduction}

Followed by the success of deep learning in various tasks, it becomes important to build a single model covering various domains without further fine-tuning to out-of-domain distribution. Because for real world application, a model is required to generalize to unseen sources of data.

In case of Question Answering (QA) task which is one of the promising areas in NLP, however, models outperforming human on SQuAD \cite{rajpurkar2016squad} cannot generalize well to other datasets. Models rather overfit to a specific dataset and require additional training on other dataset to adapt to new domain \cite{yogatama2019learning}.

Thus, in order to build a domain-agnostic QA model which is capable of handling out-of-domain data, it is necessary for model to learn domain-invariant features rather than specific ones. In this paper, we apply adversarial training framework to train a QA model with domain-agnostic representation. As shown in Figure 1, the model is divided into two components, which are the QA model and the domain discriminator. The discriminator predicts domain label of hidden representation from QA model. During the training, the QA model tries to fool the discriminator so that the hidden representation becomes indistinguishable to the discriminator. Meanwhile the discriminator is trained to identify the domain label correctly. As a result, QA model can learn domain-invariant features. Our framework can be applied to any existing QA model because the architecture of QA model stays unchanged.   

We train and validate our method on 12 datasets (6 datasets for training and 6 datasets for validation) which are provided by MRQA Shared Task. Each training dataset is considered different domain for adversarial learning in which QA model learns domain-invariant feature representation by competing with discriminator. Our experimental result shows that the proposed method improves performance compared to baseline.

% This paper is organized as follows: the preliminary of related works in section 2, detailed model description in section 3, and performance evaluation in section 4.

\begin{figure*}[h]
    \centering
    \includegraphics[width=\textwidth, height=7.5cm]{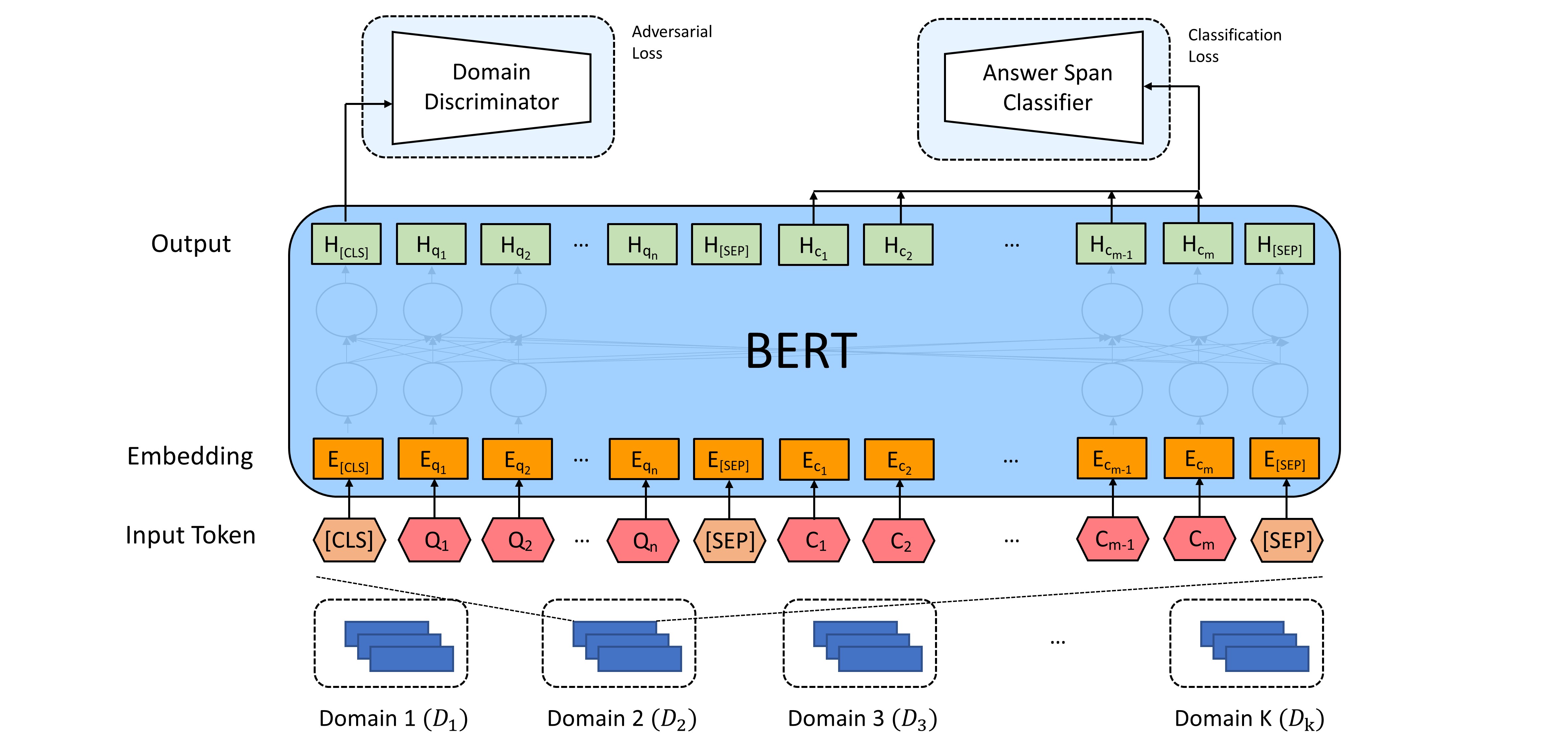}
    \caption{Overall training procedure for learning domain-invariant feature representation. Model learns to predict start and end position in the passage and fool discriminator for domain-invariant representation.}
    \label{fig:my_label}
\end{figure*}

\section{Related Works}

{\bf Pre-trained Language Model}
Recently, there have been several applications for using pre-trained language models, such as ELMo \cite{peters2018deep}, GPT \cite{radford2018improving}, or BERT \cite{devlin2018bert} to transfer the knowledge from pre-training to various downstream NLP tasks.

BERT is pretrained with bidirectional encoder \cite{vaswani2017attention} on large corpora. 
Unlike other auto-regressive language models (unidirectional or concatenation of forward and backward language model), BERT randomly masks some input tokens and predicts the masked tokens based on its context. The masked language model enables bidirectional representation, which leads to significant improvements on a number of NLP tasks, such as sentence classification, POS tagging or question answering.

{\bf Domain Generalization}
Even though many deep learning models surpass human-level performance on various task, they perform poorly on out-of-domain dataset. To address this problem, domain adaptation and domain generalization are proposed, making models more robust to out-of-domain data. The difference between domain adaptation and domain generalization is that for domain generalization, data from the target domain is not available during training.

Several methods for domain generalization exist. One of them is to train a model for each in-domain dataset. When testing on out-of-domain, select the most correlated in-domain dataset and use that model for inference \cite{xu2014exploiting}. Other works such as \cite{ghifary2015domain, muandet2013domain}, model is trained to learn a domain-invariant feature by using multi-view autoencoders and mean map embedding-based techniques.

Other approaches \cite{khosla2012undoing, li2017deeper} break down parameters of a model into domain-specific and domain-agnostic components during training with in-domain dataset, and use the domain invariant parameters for predicting data from unseen target domain.

Recently, meta-learning has been proposed for domain generalization. Some methods \cite{li2018learning, balaji2018metareg,
li2019feature} leverage meta-learning framework for domain generalization.
\newline
\newline
{\bf Adversarial Training}
The idea of adversarial training is originally proposed in the field of image generation \cite{goodfellow2014generative}, known as Generative Adversarial Network (GAN). GAN is also adopted in text generation \cite{yu2017seqgan} with policy gradient for bypassing non-differentiable operation. The concept of adversarial training is not limited to the task of generation. It can be extended to text classification \cite{ChenASWC16, DBLP:journals/corr/LiuQH17, chen-cardie-2018-multinomial}, and relation extraction \cite{wu-etal-2017-adversarial}. Likewise, attempts are made to get language-invariant features with adversarial training \cite{ChenASWC16, zhang-etal-2017-adversarial}.

Adversarial training has been used for domain adaptation or domain generalization as well. In Domain-Adversarial Neural Network (DANN) \cite{Ganin:2016:DTN:2946645.2946704}, it has two classifiers: one classifies task-specific class labels, and the other classifies whether the data belong to source or target domain. Recently, One approach \cite{Li_2018_CVPR} extends adversarial autoencoder by minimizing maximum mean discrepancy among different domains for domain-invariant feature representation.

\begin{table*}[h]
  \small
  \centering
    \begin{tabular}{p{3.4cm}cccp{6.5cm}}
        \toprule
        \textbf{Datasets} & \textbf{Samples} & \textbf{Avg.Q.len} & \textbf{Avg.P.len} & \textbf{Source} \\
        \midrule[1pt]
        \multirow{2}{*}{BioASQ (BA)} & \multirow{2}{*}{1,504} & \multirow{2}{*}{16.4} & \multirow{2}{*}{353.9} & \multirow{2}{*}{Bio-medical literature} \\
        & & & & \\
        \midrule
        \multirow{2}{*}{DROP (DP)} & \multirow{2}{*}{1,503} & \multirow{2}{*}{12.0} & \multirow{2}{*}{268.4} & \multirow{2}{*}{\parbox{6.5cm}{Wiki + National Football League (NFL) game summaries and history articles}} \\
        & & & & \\
        \midrule
        \multirow{2}{*}{DuoRC (DR)} & \multirow{2}{*}{1,501} & \multirow{2}{*}{\hspace{5pt}9.8} & \multirow{2}{*}{798.9} & \multirow{2}{*}{Wiki + IMDb} \\
        & & & & \\
        \midrule
        \multirow{2}{*}{RACE (RA)} & \multirow{2}{*}{\hspace{5pt}674} & \multirow{2}{*}{12.4} & \multirow{2}{*}{381.0} & \multirow{2}{*}{English exams for Chinese middle and high school} \\
        & & & & \\
        \midrule
        \multirow{2}{*}{RelationExtraction (RE)} & \multirow{2}{*}{2,948} & \multirow{2}{*}{11.6} & \multirow{2}{*}{\hspace{5pt}38.0} & \multirow{2}{*}{Wiki (WikiReading dataset)} \\
        & & & & \\
        \midrule
        \multirow{2}{*}{TextbookQA (TQ)} & \multirow{2}{*}{1,503} & \multirow{2}{*}{12.1} & \multirow{2}{*}{751.0} & \multirow{2}{*}{\parbox{6.5cm}{1k lessons and 26k multi-modal questions, from middle school science curriculum}} \\
        & & & & \Bstrut\\
        \bottomrule
    \end{tabular}
    \caption{Statistics of out-of-domain validation dataset. \textbf{Q} and \textbf{P} stands for question and passage, respectively. Length is calculated based on word-level token.}
  \label{tab:1}
\end{table*}

\section{Proposed Methodology}

We assume that there exists domain invariant feature representation such that QA model generalize well to predict answer on unseen out-of-domain. In order to adapt to out-of-domain, adversarial learning procedure is leveraged for learning domain-invariant representation. We present our proposed method in detail in the following sections. 

\subsection{Problem Definition}
 We formulate the task as follows: given the $K$ in-domain datasets $\mathcal{D}_i$ , consisting of triplets of passage $\mathbf{c}$, question $\mathbf{q}$, and answer $\mathbf{y}$,  where $\mathcal{D}_i = \{\mathbf{c}^{(k)}_i, \mathbf{q}^{(k)}_i, \mathbf{y}^{(k)}_i \}^{N_k}_{i=1}$. The model learned from $\{\mathcal{D}_i\}_{i=1}^{K}$  predicts answer $\mathbf{y}^l_j$ from $\mathbf{c}^l_j, \mathbf{q}^l_j$ for each $L$ out-of-domain datasets $\{ \mathcal{D}_j \}^{L}_{j=1}$.

\begin{table*}[h]
  \centering
  \small
  \begin{tabular}{l|CCCCCCCCCCCCCC}
  \toprule
    \multirow{2}{*}{\textbf{Model}}  & \multicolumn{2}{@{\hspace{1pt}}c}{\textbf{BA}} & \multicolumn{2}{@{\hspace{1pt}}c}{\textbf{DP}} & \multicolumn{2}{@{\hspace{1pt}}c}{\textbf{DR}} & \multicolumn{2}{@{\hspace{1pt}}c}{\textbf{RA}} & \multicolumn{2}{@{\hspace{1pt}}c}{\textbf{RE}} & \multicolumn{2}{@{\hspace{1pt}}c}{\textbf{TQ}} & \multicolumn{2}{@{\hspace{1pt}}c}{\textbf{Avg}} \\
        & EM & F1 & EM & F1 & EM & F1 & EM & F1 & EM & F1 & EM & F1 & EM & F1 \Tstrut\\
    \midrule[1pt]
    Bert-base & \textbf{46.44} & \textbf{60.81} & 28.31 & 37.88 & 42.78 & 53.32 & \textbf{28.23} & 39.51 & \textbf{73.33} & \textbf{83.89} & 44.30 & 52.03 & 43.90 & 54.57 \\
    Bert-base-adv & 43.35 & 60.04 & \textbf{30.51} & \textbf{40.01} & \textbf{45.97} & \textbf{57.89} & 26.50 & \textbf{39.73} & 72.67 & 83.53 & \textbf{45.62} & \textbf{55.67} & \textbf{44.10} & \hspace{2pt}\textbf{56.15} \Tstrut\\
    \midrule[1pt]
    \multirow{2}{*}{\textbf{Model}}  & \multicolumn{2}{@{\hspace{1pt}}c}{\textbf{BP}} & \multicolumn{2}{@{\hspace{1pt}}c}{\textbf{CQ}} & \multicolumn{2}{@{\hspace{1pt}}c}{\textbf{MC}} & \multicolumn{2}{@{\hspace{1pt}}c}{\textbf{MR}} & \multicolumn{2}{@{\hspace{1pt}}c}{\textbf{ST}} & \multicolumn{2}{@{\hspace{1pt}}c}{\textbf{TR}} & \multicolumn{2}{@{\hspace{1pt}}c}{\textbf{Avg}} \\
        & EM & F1 & EM & F1 & EM & F1 & EM & F1 & EM & F1 & EM & F1 & EM & F1 \Tstrut\\
    \midrule[1pt]
    Bert-base & 38.36 & 57.38 & 47.40 & 55.29 & 54.16 & 66.12 & 47.83 & 64.81 & \textbf{58.64} & \textbf{77.02} & 36.73 & 53.96 & 47.19 & 62.43 \\
    Bert-base-adv & \textbf{42.92} & \textbf{61.09} & \textbf{48.13} & \textbf{56.50} & \textbf{55.83} & \textbf{69.30} & \textbf{52.82} & \textbf{68.78} & 52.73 & 75.63 & \textbf{39.08} & \textbf{56.79} & \textbf{48.59} & \hspace{2pt}\textbf{64.68} \Tstrut\\
    \bottomrule
  \end{tabular}
  \caption{Model performance on validation and test set. Above is the validation set and below is the test set.}
  \label{tab:val_test_result}
\end{table*}

\subsection{Prediction Model}

Our method can be applied to any QA models which learn representation in the joint embedding space of passage and question. In this paper, we use BERT for QA because it is pre-trained on a large corpus and known to be generalized on several different tasks. As for standard QA task, the model is trained to minimize negative log-likelihood of answer $\mathbf{y}$ for all the given in-domain datasets, where $N, \mathbf{y}_{i,s}$, and $\mathbf{y}_{i,e}$ are respectively the total number of in-domain data, the start position and the end position of answer in the passage.
\begin{align}
\small
    \begin{split}
    \mathcal{L}_{QA} = -\frac{1}{N}\sum\limits_{k=1}^K \sum\limits_{i=1}^{N_k}& \big[ \log P_{\theta}(\mathbf{y}^{(k)}_{i, s} | \mathbf{x}^{(k)}_i, \mathbf{q}^{(k)}_i) + \\
    &\log P_{\theta}(\mathbf{y}^{(k)}_{i, e} | \mathbf{x}^{(k)}_i, \mathbf{q}^{(k)}_i) \big]
    \end{split}
\end{align}
\subsection{Adversarial Training}
\label{sect:pdf}

Minimizing the cross-entropy as in equation (1) does not ensure that the model will  generalize on unseen domain. Rather it tends to overfit to certain datasets. Inspired by GAN \cite{goodfellow2014generative}, we propose a simple yet effective method to regularize the model such that it learns domain-invariant features. 

In the adversarial training procedure, QA model learns to make the discriminator to be uncertain about its prediction. On the other hand, the discriminator is trained to classify the joint embedding of question and passage from QA model into the given $K$ domains. If the QA model can project question and passage into an embedding space where the discriminator cannot tell the difference between embeddings from different $K$ domains, we assume the QA model learns domain-invariant feature representation. 

We formulate the adversarial training as follows. A discriminator $\bm{D}$ is trained to minimize the cross-entropy loss as of equation \eqref{eq:dis}, where $l$ is domain category and $\mathbf{h} \in \mathbb{R}^{d} $ is the hidden representation of both question and passage. In our experiment, we use [CLS] token representation from BERT for $\mathbf{h}$.
\begin{align}\label{eq:dis}
    \mathcal{L}_{\bm{D}} = -\frac{1}{N}\sum\limits_{k=1}^{K} \sum\limits_{i=1}^{N_k}\log P_{\phi}(l^{(k)}_{i}|\mathbf{h}^{(k)}_i) 
\end{align}
For the QA model, it tries to maximize the entropy of $P_{\phi}(l^{(k)}_i|h^{(k)}_i)$. In other words, it minimizes Kullback-Leibler (KL) divergence between uniform distribution over $K$ classes denoted as $\mathcal{U}(l)$ and the discriminator's prediction as in equation \eqref{eq:adv}. Then the final loss for QA model is $\mathcal{L}_{QA} + \lambda \mathcal{L}_{adv}$ where $\lambda$ is a hyper-parameter for controlling the importance of the adversarial loss. In our experiments, we alternate between optimizing QA model and discriminator.
\begin{align}\label{eq:adv}
\small
\begin{split}
    \mathcal{L}_{adv} = \frac{1}{N}\sum\limits_{k=1}^K\sum\limits_{i=1}^{N_k} \infdiv{\mathcal{U}(l)}{P_{\phi}(l^{(k)}_i|\mathbf{h}^{(k)}_i)}
\end{split}
\end{align}

\section{Experiments and Result}

\subsection{Dataset}

We validate our adversarial model for MRQA Shared Task with 6 different out-of-domain datasets, which are BioASQ (BA) \cite{tsatsaronis2012bioasq}, DROP (DP) \cite{dua2019drop}, DuoRC (DR) \cite{saha2018duorc}, RACE (RA) \cite{lai2017race}, RelationExtraction (RE) \cite{levy2017zero}, and TextbookQA (TQ) \cite{kembhavi2017you}. Table 1 shows the statistics and description of these datasets. Each dataset has about 1k samples. However, the number of samples from each dataset varies. Thus, we use stratified sampling in order to make class-balanced stochastic mini-batch having certain amount of samples from all domains. We use maximum sequence length of 64 and 384 for question and passage respectively. But some examples are longer than 384. Therefore each passage is split into several chunks with a window size of 128. We discard samples without answers because all questions are considered to be answerable from given context in MRQA shared task.

Note that the final evaluation shown in the Table 2 is conducted by MRQA organizers with additional 6 out-of-domain undisclosed private test datasets, which are BioProcess (BP) \cite{scaria2013learning}, ComplexWebQuestion (CQ) \cite{talmor2018web}, MCTest (MC) \cite{richardson2013mctest}, QAMR (MR) \cite{michael2017crowdsourcing}, QAST (ST) \cite{jitkrittum-etal-2009-qast} and TREC (TR) \cite{voorhees2001trec}. 

\subsection{Implementation Details}

We implement our model based on the HuggingFace's open-source BERT implementation\footnote{https://github.com/huggingface/transformers} in Pytorch \cite{paszke2017automatic}. The performance of the baseline in our experiment differs from the official baseline of MRQA, which is based on AllenNLP \cite{Gardner2017AllenNLP}. We follow the hyperparameters as BERT for our model. In detail, we use "bert-base-uncased" with a learning rate 3e-5 and a batch size of 64. Additionally, our model requires one more hyperparameter $\lambda$, which indicates the importance of adversarial loss as described in the equation \eqref{eq:adv}. We find out that the value of 1e-2 for $\lambda$ gives the best result in our experiments. The baseline and adversarial model are trained on V100 GPU for about 5 GPU hours. For training, we use 6 in-domain datasets, which are SQuAD, TriviaQA \cite{joshi2017triviaqa}, Natural Questions \cite{kwiatkowski2019natural}, HotpotQA \cite{yang2018hotpotqa}, SearchQA \cite{dunn2017searchqa}, and NewsQA \cite{trischler2016newsqa} provided by MRQA. We select the best performing model on validation set, where models are trained for 1 or 2 epochs. The codes for our model are available at \url{https://github.com/seanie12/mrqa}.

\subsection{Performance Comparison}

Table 2 shows the performance evaluation results of models on out-of-domain datasets. In the table, the model trained with our adversarial learning is named with '-adv'. The top of the table is the result of validation datasets while the bottom is the result of test datasets. As shown in the table, overall, the model with adversarial learning has better performance compared to the baseline in terms of both EM and F1 measures. 

For validation datasets, the average F1 score of our model is about $1.5$ point higher than the baseline. In detail, our model outperforms the baseline in DP, DR, RC, and RA dataset by large margin. But the adversarial learning degrades performance in BA and RE. We can see the same aspect in terms of EM score.
Similar to the result of validation datasets, our model shows better performance in terms of EM (Exact Match) and F1 on the most of test datasets except for ST. Overall, our model has superior performance with considerable margin of over $2$ point in F1.

\section{Discussion}
In this section, we discuss some trials that have failed to improve the performance but might be helpful for future works.

\subsection{Span Refinement}
QA sample consists of a question, a passage, and  an answer span. There could exist multiple answer spans because more than one phrase in the passage can be matched with the answer text. For simplicity, only the first occurrence of answer text is used for training in most of the baseline codes. However, considering context and semantic of the given question and answer, a certain phrase in the passage is more likely to be plausible answer span relevant to the question. In order to find the most plausible answer span, a question and sentences in the passage are encoded into fixed-size vectors with universal sentence encoder \cite{cer2018universal}. We choose the span in a sentence, which is the most similar to the question in terms of cosine similarity, as golden span. In our experiment, this approach boosts up the performance of some datasets but degrades the performance a lot in the other datasets.

\subsection{Meta Learning}
We apply meta learning to domain generalization \cite{li2018learning, li2019feature, balaji2018metareg} to simulate train/test domain shift. For every epoch, one dataset is randomly selected as virtual test domain. As described in \cite{maml}, QA model is trained to maximize meta objective, which leads to improve the performance in train domain, but also in test domain. But this requires to compute Hessian-vector products, which slows down the training. This is even worse for BERT because there are 110M parameters to fine-tune. Moreover, contrary to the previous works, the meta learning for domain generalization does not help improve the performance.

\section{Conclusion}
We leverage adversarial learning to learn domain-invariant features. In our experiments, the proposed method consistently improves the performance of baseline and it is applicable to any QA model. In future work, we will try adversarial learning for pre-training model with diverse set of domains.

\section*{Acknowledgments}
We would like to thank Seonghan Ryu, Donghyeon Lee and Nicolas Remond for their valuable feedback, as well as the anonymous reviewers for their insightful comments.

\bibliography{emnlp-ijcnlp-2019}
\bibliographystyle{acl_natbib}

\end{document}